\documentclass{article}

\usepackage{microtype}
\usepackage{graphicx}
\usepackage{subfigure}
\usepackage{booktabs} 

\usepackage{hyperref}

\usepackage[accepted]{icml2023}

\usepackage{amsmath}
\usepackage{amssymb}
\usepackage{mathtools}
\usepackage{amsthm}
\usepackage{multicol}
\usepackage{multirow}

\usepackage[capitalize,noabbrev]{cleveref}

\theoremstyle{plain}

\theoremstyle{definition}

\theoremstyle{remark}

\usepackage[textsize=tiny]{todonotes}

\icmltitlerunning{Scalable Multi-Task Transfer Learning for Molecular Property Prediction}

\begin{document}

\twocolumn[
\icmltitle{Scalable Multi-Task Transfer Learning for Molecular Property Prediction}



\icmlsetsymbol{equal}{*}

\begin{icmlauthorlist}
\icmlauthor{Chanhui Lee}{lg,ku_ai}
\icmlauthor{Dae-Woong Jeong}{lg}
\icmlauthor{Sung Moon Ko}{lg}
\icmlauthor{Sumin Lee}{lg}
\icmlauthor{Hyunseung Kim}{lg}
\icmlauthor{Soorin Yim}{lg}
\icmlauthor{Sehui Han}{lg}
\icmlauthor{Sungwoong Kim}{ku_ai}
\icmlauthor{Sungbin Lim}{lg,ku_stat}
\end{icmlauthorlist}


\icmlaffiliation{lg}{LG AI Research, Seoul, Republic of Korea}
\icmlaffiliation{ku_ai}{Department of Artificial Intelligence, Korea University, Seoul, Republic of Korea}
\icmlaffiliation{ku_stat}{Department of Statistics, Korea University, Seoul, Republic of Korea}

\icmlcorrespondingauthor{Sungbin Lim}{sungbin@korea.ac.kr}

\icmlkeywords{Molecular property prediction; Multi-task learning; Transfer learning}


\vskip 0.3in
]



\printAffiliationsAndNotice{} 


\begin{abstract}

Molecules have a number of distinct properties whose importance and application vary.
Often, in reality, labels for some properties are hard to achieve despite their practical importance. 
A common solution to such data scarcity is to use models of good generalization with transfer learning. 
This involves domain experts for designing source and target tasks whose features are shared. 
However, this approach has limitations: i). Difficulty in accurate design of source-target task pairs due to the large number of tasks, and ii). corresponding computational burden verifying many trials and errors of transfer learning design, thereby iii). constraining the potential of foundation modeling of multi-task molecular property prediction.
We address the limitations of the manual design of transfer learning via data-driven bi-level optimization. 
The proposed method enables scalable multi-task transfer learning for molecular property prediction by automatically obtaining the optimal transfer ratios.
Empirically, the proposed method improved the prediction performance of 40 molecular properties and accelerated training convergence.

\end{abstract}

\section{Introduction}
\label{sec:intro}
Given that a molecule has a number of molecular properties, basically, molecular property prediction is to predict a target property among various properties \cite{Wieder2020ACR}.
There are many important applications of molecular property prediction, including virtual screening and discovery of novel materials and drugs \cite{Christopher2001ExperimentalAC, Atanasov2021NaturalPI, Gentile2022ArtificialIV, Sadybekov2023ComputationalAS}.
Molecular property prediction provides vital information in making informed decisions throughout the discovery and development process so that the development cycle of the product can be accelerated.

However, in reality, molecular property prediction often suffers from the data scarcity problem \cite{Hu2019PretrainingGN, Li2022KPGTKP, Li2021GeomGCLGG} due to various factors, including the high cost of experimental data generation \cite{Axelrod2022GEOMEM}, the complexity of chemical compounds \cite{Kumar2020ARO}, and the proprietary nature of pharmaceutical data \cite{Heyndrickx2023MELLODDYCF}. 
As a result, researchers have tried to overcome this limitation, such as using advanced computational techniques like transfer learning \cite{ko2024geometrically}, data augmentation \cite{You2020GraphCL}, and other approaches \cite{Lu2019MolecularPP, Yao2023EnhancingMP, Qian2023CanLL} that can learn effectively from smaller datasets.

Transfer learning \cite{Pan2010ASO, Zhuang2019ACS} allows for the effective generalization of knowledge learned from source task data distribution to the target task data.
Effective transfer can be fulfilled through learning mutually informative feature representations between aligned tasks.
GATE \cite{ko2024geometrically} introduced a geometric alignment of various tasks to enhance task alignment for molecular property prediction.
With the proposed geometrical alignment, the prediction model can learn geometrically aligned molecular representations that are applicable from source task to target task, enabling effective transfer learning and surpassing the performance of baseline models.

Recently, \citet{ko2024multitask} extended GATE to a multi-task setting by sharing a single latent space among multiple tasks, applying geometrical alignment regularization within this shared latent space.
In this extended formulation of GATE, different transfer ratios can be applied for each (source, target) task pair, which represent a belief in how much a source task can be helpful to the target task. 
\begin{figure}[h!]
  \centering
  \includegraphics[width=0.4\textwidth]{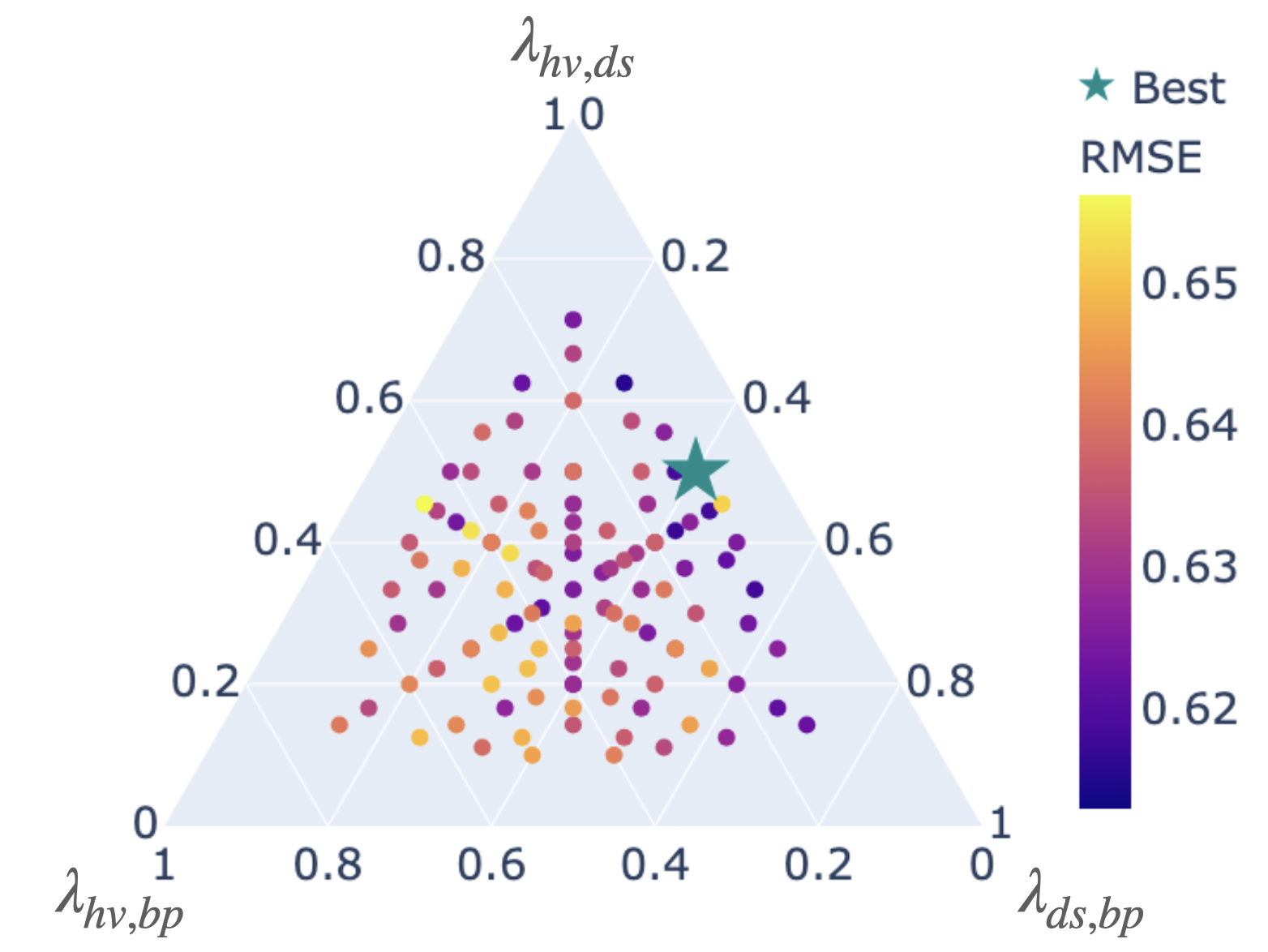}
  \caption{
  Grid search of transfer ratios $\lambda$ between density (ds), heat of vaporization (hv), and boiling point (bp).
  Each axis $\lambda_{hv, bp}, \lambda_{ds, bp}, \lambda_{hv,ds}$ corresponds to the transfer ratio between (hv, bp), (ds, bp), (hv, ds), assuming $\lambda_{i \rightarrow j}=\lambda_{j \rightarrow i}$.
  The color of a point corresponds to the Root Mean Square Error (RMSE) of model prediction at the end of training with $\lambda_{hv, bp}, \lambda_{ds, bp}, \lambda_{hv, ds}$, and the best hyperparameter set is marked as a star.
  }
  \label{fig:grid}
\end{figure}
In \Cref{fig:grid}, we investigated the effect of different transfer ratios for three molecular properties: density (ds), heat of vaporization (hv), and boiling point (bp).
Assuming the same transfer ratio between paired tasks, we conducted a grid search over [0.2, 1.0] for 3 hyperparameters $\lambda_{hv, bp}, \lambda_{ds, bp}, \lambda_{hv, ds}$ and found that overall prediction performance largely varies by the setting of transfer ratios.

Though the transfer ratio could have a large effect on the prediction performance, GATE lacks a method for exploring transfer ratios but rather uses them as hyperparameters, which presents several limitations.
i. Inaccuracy in predicted transfer ratios by domain experts: Given the black-box nature of deep learning models, there is no guarantee that the source task and the corresponding source data chosen by a domain expert will actually enhance the performance of the target task.
ii. Limited scalability: As the number of tasks increases, manually setting proper ratios for all possible transfer interactions between two tasks by a domain expert becomes infeasible and less optimized.
iii. Constraining the foundation modeling in molecular property prediction: As different tasks in molecular property prediction fundamentally involve comprehending molecular structures, the performance of tasks with limited data can be improved by merging existing datasets for extensive multi-task training. 
This approach maximizes the benefits of foundational modeling in predicting molecular properties.

To address these limitations, we propose a novel bi-level optimization method to automatically obtain the optimal transfer ratios for given multi-task data.
This bi-level optimization replaces previous manual hyperparameter searches by domain experts through gradient-based learning on the validation performance.
The training algorithm remains the same during the training phase; the difference occurs in the validation phase.
In the validation phase, gradients flow from the computed loss to the computation node representing the transfer ratio and are updated gradient-based on their contribution to the loss value during the validation phase.
Since the gradient computation is restricted to the transfer ratio, additional time and space costs for the proposed bi-level optimization are negligible.
The proposed gradient-based bi-level optimization efficiently obtains the optimal transfer ratio, especially on a large task space, without cumbersome tuning by domain experts.

\paragraph {Contribution}
\begin{itemize}
    \item We propose a data-driven method to search optimal transfer ratios for multi-task transfer learning of molecular property prediction.
    \item The proposed method has improved the performances on 40 tasks of molecular property regression.
    \item The proposed method accelerates the convergence of multi-task transfer learning in molecular property regressions.
\end{itemize}

\section{Preliminary: Multi-task property regression}
\label{sec:method}
This section introduces preliminary works for multi-task property regression.
For multi-task transfer learning in property regression, we leverage GATE algorithms extended for multi-task transfer learning \cite{ko2024multitask}.

\subsection{Multi-task learning extension of GATE}
GATE addresses transfer learning among a number of tasks, introducing additional side tasks to learn mutually useful features for different tasks in shared manifold $\mathcal{M}$.
$\mathcal{M}$ is a manifold where each task-specific model can learn the general geometrical knowledge of molecular structure.
This strategy guides the model in learning generally useful features for molecular property regression, allowing the task of scarce data to take advantage of knowledge learned from another data-enriched task.

\begin{figure}[h!]
  \centering
  \includegraphics[width=0.5\textwidth]{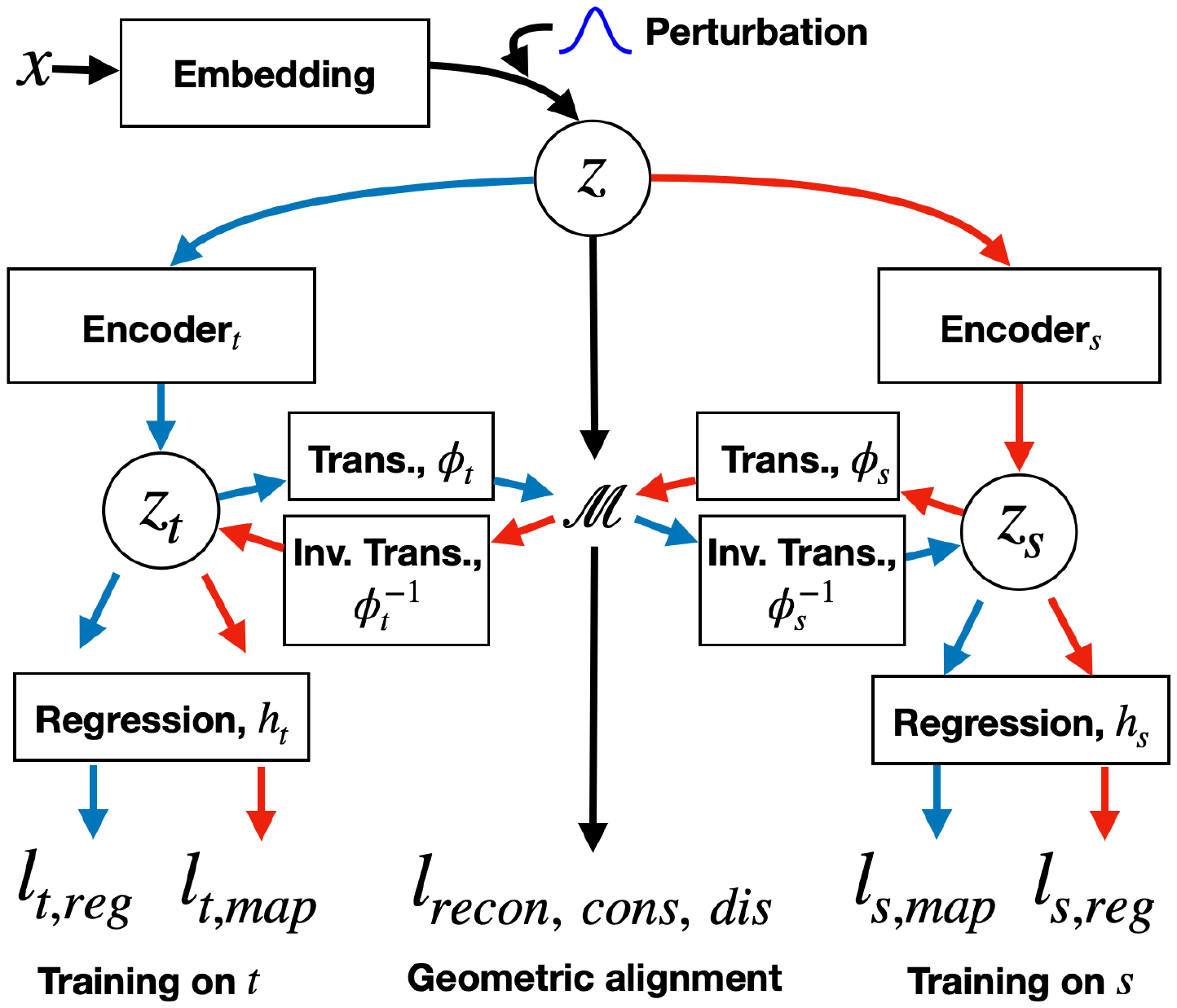}
  \caption{
  Training overview of GATE. $t,s$ represents the target and source tasks for transfer learning.
  The colors of the arrows differentiate prediction paths: red corresponds to the path from $\text{Encoder}_s$, and blue corresponds to the path from $\text{Encoder}_t$.
  }
  \label{fig:overview}
\end{figure}

\label{sec:method-multi}

\subsection{Target task regression}
When a molecule $x$ is fed into an embedding model as SMILES \cite{Weininger1988SMILESAC}, we get a corresponding embedding vector $z$.
Given a target task $t$, a task-specific encoder ($\text{encoder}_t$) embeds $z$ into a latent vector $z_t$ on the manifold for the target task $t$.
Then, the regression head $h_t$ predicts $\hat{y}_t$ to calculate the Mean Squared Error (MSE) loss with respect to target label $y_t$.
\begin{gather}
    z_t=\text{encoder}_{t}(z) \\
    \hat{y}_t=h_t(z_t) \\
    l_{\mathrm{reg}} = \frac{1}{N}\sum_{i}^{N}\mathrm{MSE}(y_t, \hat{y}_{t})
\end{gather}
where $N$ is the number of data points.

\subsection{Transfer learning from source task}
Let $s$ be another task we can leverage for target task learning.
Thanks to the shared manifold $\mathcal{M}$, $z_t$ can be represented from a source task representation $z_s$, via transformation $\phi_{s \rightarrow \mathcal{M}}$ and inverse transformation $\phi^{-1}_{\mathcal{M}\rightarrow t}$.
\begin{gather} 
\label{transform1}
    z_s = \text{encoder}_{s}(z) \\
    z_\mathcal{M} = \phi_{s\rightarrow \mathcal{M}}(z_s) \\ 
    z_t = \phi^{-1}_{\mathcal{M} \rightarrow t}(z_\mathcal{M}) \\
    \hat{y_t} = h_t(z_t)
\end{gather}
where $\phi_{s\rightarrow \mathcal{M}}(z_s)$ means a transformation of the vector from the manifold of source task $s$ to shared manifold $\mathcal{M}$, and $\phi^{-1}_{\mathcal{M} \rightarrow t}(z_s)$ means an inverse transformation from the shared manifold $\mathcal{M}$ to the manifold of target task $t$. 

With a hyperparameter called a mapping ratio $\lambda_{s \rightarrow t}$, GATE conducts transfer learning from source task $s$ to target task $t$.
\begin{equation}
\label{loss_map}
    l_{map} = \sum_s \frac{1}{N}\sum_{i}^{N}\lambda_{s \rightarrow t}\mathrm{MSE}(y_t, \hat{y}_{t})
\end{equation}
As the correlation varies across different source-target task pairs, the effectiveness of multi-task transfer learning is dependent on the proper search of the $\lambda_{s \rightarrow t}$.
For instance, the Highest Occupied Molecular Orbital (HOMO) and Lowest Unoccupied Molecular Orbital (LUMO) tasks would be deeply correlated, sharing many of the necessary features representing molecular orbitals. 
Therefore, a high $\lambda_{s \rightarrow t}$ value can accelerate the mutual learning of the HOMO and LUMO.
However, in a multi-task learning setting, the correlation of the target-source task is conditioned on other tasks, which makes it more difficult to find optimal $\lambda$ with many tasks to learn.
In this situation, completing the entire correlation of target-source task pairs is prohibitive even for experienced domain experts.

\subsection{Geometric regularizations}
To align manifolds of different tasks to be the shared manifold $\mathcal{M}$, GATE aims to align the geometric representation of a molecule in different properties through side tasks: reconstruction, consistency, and distance.

\paragraph{Reconstruction} 
To convince those models can learn general geometries useful across tasks, $\mathcal{M}$ should have enough expressiveness to reconstruct $z$.
The following reconstruction loss
\begin{equation}
    l_{\mathrm{ae}} = \sum_i \mathrm{MSE}(z_i, \phi_{\mathcal{M} \rightarrow i}(\phi^{-1}_{i \rightarrow \mathcal{M}}(z_i)),
\end{equation}
regularizes GATE to maintain the reconstruction capability.

\paragraph{Consistency}
Too much divergence between the manifold of $t, s$ could harm the transfer effect.
To regularize the significant divergence between $z_t, z_s$ on the shared manifold $\mathcal{M}$, GATE introduces a loss for consistency,

\begin{equation}
    l_{cons} = \sum_s \mathrm{MSE}(z_s, z_t).
\end{equation}

\paragraph{Distance}

To learn robust transformation $\phi_{i \rightarrow \mathcal{M}}, \phi^{-1}_{\mathcal{M} \rightarrow i}$ under perturbation, GATE applies a perturbation on $z$ to have $z'$.
Then regularizes the resulting displacement for different tasks $i$ to be minimized as

\begin{equation}
    l_{dis} = \frac{1}{M}\sum_s C_s\sum_{p=1}^M \mathrm{MSE}(s^p_s, s^p_t),
\end{equation}
where $M$ is the number of perturbed points, superscript $p$ means $p$-th perturbed point, $C_s$ is the hyperparameter, and $s^p_s=\|\phi_{i \rightarrow \mathcal{M}}(\text{encoder}_i(z)) - \phi_{i \rightarrow \mathcal{M}}(\text{encoder}_i(z^p))\|$.

Aggregating the losses for geometrical alignments, the total loss is calculated as
\begin{equation}
    l_{tot} = l_{reg} + l_{ae} + l_{cons} + l_{dis} + l_{map},
\end{equation}
to update model parameters for $\phi, \phi^{-1}, h$, and the embedding model.
For brevity, we simply represent the corresponding model parameters as $\theta$.

\section{Bi-level optimization of GATE}
We interpret the problem of finding optimal $\lambda$ as a bi-level optimization problem:
\begin{equation}
\begin{aligned}
    &\underset{\lambda}{\min} \quad l_{\text{val}}(\theta, \lambda) \\
    &\text{s.t.}\quad \theta^*(\lambda)=\underset{\theta}{\arg\min} \ l_{\text{train}}(\theta, \lambda),
\end{aligned}
\end{equation}
\Cref{alg:biopt} depicts the detailed algorithms for the bi-level optimization of GATE.
First,  $\theta$ is updated using the training dataset $D_{tr}$ and $\lambda$ in the inner loop, and in the outer loop, based on the updated $\theta$, $\lambda$ is updated with respect to the performance in the validation dataset $D_{val}$.
\begin{algorithm}[h!]
   \caption{Bi-level Optimization for GATE}
   \label{alg:biopt}
\begin{algorithmic}[1]
   \STATE {\bfseries Input:} Training data $D_{tr}$, validation data $D_{val}$
   \STATE Initialize model $\theta$, transfer ratio $\lambda$, transfer momentum $m, v$
   \REPEAT
    \STATE $\theta \leftarrow \underset{\theta}{\arg \min} \ L(D_{tr}, \theta, \lambda)$; {Inner loop}
    \STATE $\lambda \leftarrow \underset{\lambda}{\arg \min} \ L(D_{val}, \theta, \lambda, m, v)$; {Outer loop}
   \UNTIL{converged}
\end{algorithmic}
\end{algorithm}

In the inner loop, our objective is to update $\theta$ given $\lambda$ and $D_{tr}$.
Given a molecule, $x$, the embedding model projects $x$ into the embedding vector $z$. 
Subsequently, $\text{encoder}_i$ projects $z$ to $z_i$ for all tasks $i \in T \cup S$, where $T, S$ represent sets of target tasks and source tasks, respectively.
Then, we can get two predictions for the target property $t$; one is directly from $z_t$ and the other is from $\phi^{-1}_{\mathcal{M} \rightarrow t}(\phi_{s \rightarrow \mathcal{M}}(z_s))$.
The predicted values are used to calculate the regression loss $l_{reg}$ and mapping loss $l_{map}$, respectively.
After summation of losses $l_{reg}$ and $l_{map}$ with $l_\text{ae}, l_\text{cons}, l_\text{dis}$, finally $\theta$ is updated from the corresponding gradient.

\begin{algorithm}[h!]
   \caption{Inner loop}
   \label{alg:inner}
\begin{algorithmic}[1]
   \STATE {\bfseries Input:} Target tasks $T=\{t_1, \cdots, t_{N_T}\}$, Source tasks $S=\{s_1, \cdots, s_{N_S}\}$, model parameters $\theta$, transfer ratios $\lambda$, training data ($x_i, y_i$) $\forall i \in T \cup S$
   \STATE {\bfseries Output:} Optimized model parameters $\mathbf{\theta}$
   \WHILE{Training epoch}
    \STATE $l_{tot}=0$
    \STATE $z=\text{embedding}(x),\ \forall i \in T \cup S$ 
    \STATE $z_i=\text{encoder}_i(z),\ \forall i \in T \cup S$ 
    \FOR{$(t,s)$ \bfseries{in} $\{(t, s)| t \in T, s \in S\}$}
    \STATE $l_{reg} =\mathrm{MSE}(y_t, h_t(z_t))$
    \STATE $l_{map} =\lambda_{s \rightarrow t} \mathrm{MSE}(y_t, h_t(\phi^{-1}_{\mathcal{M} \rightarrow t}(\phi_{s \rightarrow \mathcal{M}}(z_s))))$
    \STATE $l_{tot} \mathrel{+}= l_{reg} + l_{map} + l_{ae} + l_{cons} + l_{dis}$
    \ENDFOR
    \STATE $\theta \leftarrow \nabla_\theta l_{tot}$
    \ENDWHILE
\end{algorithmic}
\end{algorithm}

In the outer loop, we aim to search for $\lambda=\underset{\lambda}{\min} \quad L_{\text{val}}(\theta^*, \lambda)$. 
The mapping loss between the target and source task is calculated to get gradient $g$ with respect to the updated $\theta$ through the inner loop.
Then $g$ is used to update the moving average $m$ and the squared moving average $v$, with the corresponding hyperparameters $\beta_0, \beta_1 \in [0, 1)$.
Finally, the bias-corrected estimate of the first moment and the second moment, $\hat{m}$ and $\hat{v}$, are calculated to update the transfer ratio $\lambda$.
Markedly, the calculated mapping loss in the outer loop only updates $\lambda$ without affecting $\theta$.
This procedure is a data-driven hyperparameter search that substitutes for manual hyperparameter search, which is a bottleneck for the foundation modeling of multi-task transfer learning for molecular property regression.

\begin{algorithm}[h!]
   \caption{Outer loop}
   \label{alg:outer}
\begin{algorithmic}[1]
   \STATE {\bfseries Input:} Target tasks $T=\{t_1, \cdots, t_{N_T}\}$, Source tasks $S=\{s_1, \cdots, s_{N_S}\}$, model parameters $\theta$, transfer ratio $\lambda$, transfer momentum $m, v$, validation data ($x_i, y_i$) $\forall i \in T \cup S$
   \STATE {\bfseries Output:} Optimized $\lambda$
   \WHILE{Validation epoch}
    \STATE $\text{step} \leftarrow \text{step} + 1$
    \STATE $l_{map}=0$
    \STATE $z=\text{embedding}(x),\ \forall i \in T \cup S$ 
    \STATE $z_i=\text{encoder}_i(z),\ \forall i \in T \cup S$ 
    \FOR{$(t,s)$ \bfseries{in} $\{(t, s)| t \in T, s \in S\}$}
    \STATE $l_{map} \mathrel{+}=\lambda_s \mathrm{MSE}(y_t, h_t(\phi^{-1}_{\mathcal{M} \rightarrow t}(\phi_{s \rightarrow \mathcal{M}}(z_s))))$
    \ENDFOR
    \STATE $g \leftarrow \nabla_\lambda l_{map}$
    \STATE $m\leftarrow \beta_0m + (1 - \beta_0)g,\ v\leftarrow \beta_1v + (1 - \beta_1)g^2$
    \STATE $\hat{m} \leftarrow \frac{m} {1 - \beta_0^{\text{step}}},\ \hat{v} \leftarrow \frac{v} {1 - \beta_1^{\text{step}}}$
    \STATE $\lambda \leftarrow \lambda - \eta\frac{\hat{m}} {\sqrt{\hat{v}} + \epsilon}$
    \ENDWHILE
\end{algorithmic}
\end{algorithm}

\section{Experiments}
\label{sec:exp}

\subsection{Dataset}
\label{sec:exp-dataset}

To test in a scaled-up multi-task setting, we collected data from 40 tasks from PubChem \citep{10.1093/nar/gkac956}, Ochem \citep{sushko2011online}, CCDDS, Yaws Handbook, and Jean-Claude Bradley.
We specified each task and number of data points in \Cref{appx:dataset}.
For the robust test, we used scaffold split of the train and test dataset based on the molecular structure \citep{bemis1996properties}.
To avoid the overfitting of transfer ratio adaptation to the validation dataset, we interchanged 20\% of the train and validation datasets for every epoch.

\subsection{Results}
\label{sec:exp-results}

\begin{table*}[h!]
\caption{
40 task RMSE of GATE and GATE integrated with bi-level optimization (GATE*).
The best case for each task is highlighted in bold.
}
\centering
\begin{small}
\begin{sc}
\begin{tabular}{l|cccccccccc}
\toprule
 & \multicolumn{10}{c}{Tasks} \\
Method & cps & lm & uft & vs & mas & ctp & gef & pka & mvs & dm \\ \midrule
GATE    & 0.154 & 0.367 & 0.497 & \textbf{0.650} & \textbf{0.343} & 0.402 & 0.326 & 0.765 & 0.250 & 0.690 \\
GATE*   & \textbf{0.139} & \textbf{0.366} & \textbf{0.475} & 0.739 & 0.376 & \textbf{0.373} & \textbf{0.236} & \textbf{0.731} & \textbf{0.233} & \textbf{0.687} \\\midrule
 & sef & cpl & hvc & fp & par & ct & vp & lp & cpg & ip \\\midrule
GATE    & 0.452 & 0.252 & 0.562 & 0.645 & \textbf{0.241} & 0.414 & 0.730 & \textbf{0.134} & \textbf{0.119} & \textbf{0.501} \\
GATE*   & \textbf{0.358} & \textbf{0.133} & \textbf{0.555} & \textbf{0.636} & 0.281 & \textbf{0.408} & \textbf{0.685} & 0.135 & 0.134 & 0.533 \\\midrule
 & spa & lf & st & hf & ri & as & ds & nec & dk & mvl \\\midrule
GATE    & 0.259 & \textbf{0.510} & \textbf{0.409} & 0.561 & 0.443 & \textbf{0.754} & 0.477 & 0.323 & 0.591 & 0.484 \\
GATE*   & \textbf{0.234} & 0.512 & 0.420 & \textbf{0.516} & \textbf{0.418} & 0.807 & \textbf{0.442} & \textbf{0.257} & \textbf{0.559} & \textbf{0.444} \\\midrule
 & hv & sae & bp & ctv & mp & hc & pol & hm & aw & rog \\\midrule
GATE    & 0.478 & 0.263 & 0.520 & 0.284 & 0.583 & 0.265 & 0.307 & 0.441 & 0.711 & 0.552 \\
GATE*   & \textbf{0.431} & \textbf{0.258} & \textbf{0.508} & \textbf{0.237} & \textbf{0.579} & \textbf{0.246} & \textbf{0.285} & \textbf{0.432} & \textbf{0.596} & \textbf{0.548} \\
\bottomrule
\end{tabular}
\end{sc}
\end{small}
\label{tab:rmse}
\end{table*}

\Cref{tab:rmse} shows the performance of GATE with and without bi-level optimization on 40 different molecular property regression tasks. 
To evaluate the effectiveness of the proposed method, other than $\lambda$, we used the same model architecture and hyperparameters of GATE.
Performance is measured in terms of Root Mean Square Error (RMSE), a standard metric used to measure prediction accuracy. A lower RMSE indicates better performance.
The results show that GATE with bi-level optimization generally achieves lower RMSE scores across most tasks than the original GATE method. 
This is achieved by learning the transfer ratio $\lambda$ to minimize prediction error described in \Cref{loss_map}, which enables more effective transfer learning than using constant $\lambda$, though given the same data for target task and source tasks.
Specifically, as \Cref{tab:win_count}, the performances were enhanced in 31 out of 40 tasks, reducing the average RMSE by 4.4\%.
This suggests that incorporating bi-level optimization in GATE improves prediction accuracy across a wide range of tasks. 

\begin{table}[h!]
\label{tab:win_count}
\caption{40 task RMSE Improvements by applying bi-level optimization over vanilla GATE}
\vskip 0.15in
\begin{center}
\begin{small}
\begin{sc}
\begin{tabular}{lcccr}
\toprule
Method & No. Improved Tasks & Avg RMSE \\
\midrule
GATE    & -  & 100\% \\
GATE*   & 31 & 95.6\%\\

\bottomrule
\end{tabular}
\end{sc}
\end{small}
\end{center}
\vskip -0.1in
\end{table}

\begin{figure}[h!]
  \centering
  \includegraphics[width=0.95\columnwidth]{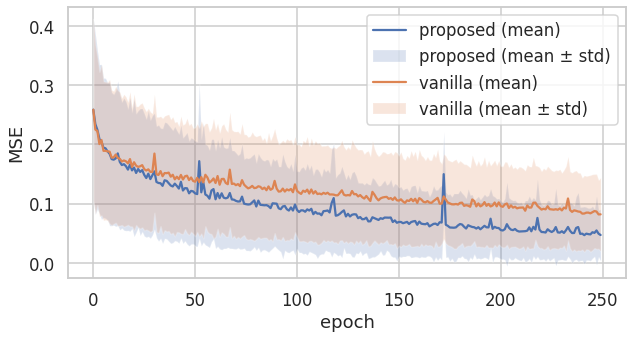}
  \caption{
  Validation loss curve in learning 40 tasks molecular property regression, with and without the proposed methods.
  }
  \label{fig:val_curve}
\end{figure}
In addition, we found that applying bi-level optimization not only enhances performance but also accelerates loss convergence.
\Cref{fig:val_curve} shows that with the same training epoch, regression loss converges much faster with the proposed method.
The fast convergence is due to the learning of $\lambda_{s \rightarrow t}$, which strengthens the transfer between highly correlated molecular properties and, at the same time, controls too much transfer between less correlated molecular properties.
In the end, the variance across the different tasks is reduced, as shown by the narrower shade of the proposed method than vanilla GATE.
\section{Discussion}
\Cref{alg:outer} imply that we can accelerate the outer loop with reduced GPU memory usage by only backpropagating the gradient of tasks whose $\lambda$ is above a threshold.
Training 40 molecular property prediction tasks, we found that this direction is promising, as the 95\% of quadratic variations of $\lambda$ are under 0.1.
This means that updates from many source and target task pairs do not result in a big update of the model parameters.
We hope this direction guides future works to improve the time and space complexity of the proposed method.

\section{Conclusion}
\label{sec:conclusion}
This study presents a bi-level optimization approach for enhancing transfer learning in multi-task property regression on a large scale. 
The performance in multi-task transfer learning is significantly influenced by how the correlation between the source and target tasks is modeled. 
Typically, designing this correlation has relied on domain experts. 
However, with increasing tasks, relying solely on domain experts for correlation design needs impractical time and inaccurate design due to the exponentially increasing number of task pair combinations, which can lead to sub-optimal outcomes. 
To address this issue, we employ a data-driven bi-level optimization strategy to identify the optimal correlation design. 
In our evaluation across 40 tasks, applying our method decreased RMSE for 31 tasks, with an average reduction of 4.4\%.


\section*{Acknowledgements}
This work was supported by LG AI Research.
This work was also supported by Institute of Information \& communications Technology Planning \& Evaluation(IITP) grant funded by the Korea government(MSIT)(No. 2022-0-00612, Geometric and Physical Commonsense Reasoning based Behavior Intelligence for Embodied AI; No. 2019-0-00079, Artificial Intelligence Graduate School Program, Korea University), and National Research Foundation of Korea(NRF) funded by the Korea government(MSIT)(2021R1C1C1009256), and a grant from Korea University (K2407521).

\bibliography{example_paper}
\bibliographystyle{icml2023}

\newpage
\appendix
\onecolumn
\section{Dataset}
\label{appx:dataset}
In this section, we provide the task set used for 40 task molecular property prediction, with their full name and corresponding number of data points.

\begin{table*}[h!]
\caption{
Dataset configuration used for the experiment of 40 tasks property prediction.
}
\centering
\begin{small}
\begin{sc}
\begin{tabular}{l|c|c}
\toprule
Property & Abbreviation & \# data points \\
\midrule
Heat of vaporization & hv & 1504 \\
Viscosity & vs & 1307 \\
Surface tension & st & 977 \\
Density & ds & 3079 \\
Boiling point & bp & 8044 \\
Refractive index & ri & 11143 \\
Melting point & mp & 22901 \\
LogP & lp & 28268 \\
Abraham descriptor S & as & 1915 \\
Dielectric constant & dk & 999 \\
Dipole moment & dm & 11224 \\
Flash point & fp & 9409 \\
Ionization potential & ip & 898 \\
pKa & pka & 9514 \\
Polarizability & pol & 457 \\
Vapor pressure & vp & 4262 \\
Absorbance maximum wavelength & aw & 11896 \\
Critical temperature & ct & 2414 \\
Heat of combustion & hc & 2118 \\
Hydration free energy & hf & 648 \\
Lower flammability limit temperature & lf & 1646 \\
HOMO energy level & hm & 97262 \\
LUMO energy level & lm & 97262 \\
Molar heat capacity (liquid) & cpl & 387 \\
Molar heat capacity (gas) & cpg & 264 \\
Molar heat capacity (solid) & cps & 218 \\
Molar volume (liquid) & mvl & 8513 \\
Molar volume (solid) & mvs & 218 \\
Heat of vaporization & hvc & 1957 \\
Critical pressure & ctp & 3007 \\
Critical volume & ctv & 2413 \\
Gibbs Energy of Formation for Ideal gas & gef & 1828 \\
Magnetic Susceptibility & mas & 432 \\
Net Standard State Enthalpy of Combustion & nec & 1182 \\
Parachor & par & 960 \\
Radius of Gyration & rog & 1370 \\
Solubility Parameter& spa & 1509 \\
Standard State Absolute Entropy & sae & 1072 \\
Standard State Enthalpy of Formation & sef & 1638 \\
Upper Flammability Limit Temperature & uft & 1443 \\
\bottomrule
\end{tabular}
\end{sc}
\end{small}
\label{tab:dataset}
\end{table*}

\end{document}